\documentclass[11pt,a4paper]{article}

\usepackage[utf8]{inputenc}
\usepackage[T1]{fontenc}
\usepackage{lmodern}
\usepackage[margin=1in]{geometry}
\usepackage{microtype}
\usepackage{needspace} 
\usepackage{amsmath,amssymb,amsthm}
\usepackage{graphicx}
\usepackage{listings}
\usepackage{xcolor}
\usepackage{booktabs}
\usepackage{array}
\newcolumntype{L}[1]{>{\raggedright\arraybackslash}p{#1}}
\usepackage{tikz}
\usetikzlibrary{arrows.meta,positioning}
\usepackage{natbib}
\usepackage[colorlinks=true,linkcolor=blue!60!black,citecolor=blue!60!black,urlcolor=blue!60!black]{hyperref}
\usepackage{doi} 

\newtheorem{example}{Example}

\definecolor{celllabel}{RGB}{96,110,150}
\lstdefinestyle{wolfram}{
    language=Mathematica,
    basicstyle=\ttfamily\small,
    keywordstyle=\color{blue!70!black},
    commentstyle=\color{gray},
    stringstyle=\color{orange!80!black},
    breaklines=true,
    frame=single,
    columns=fullflexible,
    keepspaces=true,
    showstringspaces=false,
    escapeinside={(*@}{@*)}
}
\newcommand{\IN}[1]{{\ttfamily\small\color{celllabel}In[#1]:=\ }}
\newcommand{\OUT}[1]{{\ttfamily\small\color{celllabel}Out[#1]=\ }}

\title{Implementing Computational Law in Wolfram Language for the Governance of Artificial Intelligence}
\author{James K. Wiles\\
\textit{Wolfram Institute for Computational Foundations of Science}}
\date{}

\begin{document}

\maketitle

\begin{abstract}
How do we govern AI systems whose reasoning we cannot fully inspect? Governance does not require understanding a system's reasoning. It requires stating what the system is obliged, permitted, and forbidden to do, and checking whether it complied. I present an implementation of Reified Input/Output Logic, the formalism behind the DAPRECO knowledge base, in Wolfram Language: the core I/O axioms, obligations, permissions, constitutive norms, reified eventualities, and temporal operators. I then test whether GPT-4 can translate English legal statements into the formalism, and report the failures: hallucinated functions, omitted temporal scope, deviation from the formalism, and (in the worst cases) code that runs, reads plausibly, but silently encodes the wrong norm. A case study, an AI guard dog operating under a computational contract, shows how formalized rules can extend from a contract directly into the operational code of an embodied agent, producing symbolic, auditable justifications for its behaviour. I argue that computational law can be used as a governance tool and that a desirable goal would be to formalize the law that can be formalized and ought to be programmatically executable.

\medskip
\noindent\textbf{Keywords:} computational law, input/output logic, reification, AI governance, Wolfram Language, large language models, deontic logic
\end{abstract}

\section{Introduction}
\label{sec:introduction}

``Fire the nukes!'' says the Government AI. ``Why?!'' you ask, as President of Mars. The Chief Machine Officer explains: ``It's a black box, sir. We have no idea. But in simulation it does what we think is right 99\% of the time.''

Now consider an alternate briefing: ``It's a black box, sir, but it has a computational law obligation to defend us at all costs, which it adheres to 99\% of the time; otherwise it gets digitally executed.''

Which future would you rather live in?

The thought experiment is provocative but the choice is real. We already trust black-box systems with loan approvals, medical diagnoses, and content moderation, and this is moving toward increasingly important decisions. Interpretability research may one day let us see inside these systems, and understand why they conclude certain things, but governance does not have to wait for that. To govern a system, we need to state what it is obliged, permitted, and forbidden to do, and we need to check whether it complied. Humans already run institutions that do exactly this: legal systems. Their rules, the obligations they impose, and the penalties they attach are all written down in legal documents.

The way law is encoded is the problem: law is written in human language. Human language tries to capture the full experience of our world, and ambiguity is part of how it does so \citep{piantadosi2012communicative}. Law inherits that ambiguity because it has no other medium. Statutes, contracts, and judgments are human-language text, and a dispute over what they mean is a dispute over language. Large language models now read and write legal text fluently, but they interpret it the way humans do: opaquely. In legal reasoning, where the position of a comma can change the meaning of a document, ambiguity can cause harm. \emph{Computational law}, the formalization of legal reasoning into executable code \citep{genesereth2015computational,genesereth2021what}, offers a different route. Instead of hoping AI systems interpret our written laws correctly, write laws that AI systems can directly execute.

This paper is an exploratory implementation study, not a systematic evaluation. It contributes four things. First, an implementation of the core of Reified Input/Output Logic \citep{robaldo2017reification}, the formalism behind the DAPRECO knowledge base \citep{robaldo2020dapreco}, in Wolfram Language. Second, a test of whether GPT-4 can translate English legal statements into that implementation, reported in full, wrong outputs included. The failures are instructive: three of the four translations produced code that ran and looked correct while silently encoding the wrong norm. Two imposed obligations on the wrong people, one of them while stating the correct expectation in its own comments, and one measured legal compliance with the wrong geometry. Third, a case study in which a computational contract extends into the operational code of an AI agent, a robotic guard dog named Fluffy, producing symbolic, auditable justifications for its actions. Fourth, an argument about what any of this buys AI governance, resting on two distinctions that often get muddled: governance \emph{of} AI versus governance \emph{in} AI, and governance versus ethics.

Section~\ref{sec:background} covers the background. Section~\ref{sec:implementation} presents the implementation. Section~\ref{sec:llm-bridge} reports the GPT-4 experiments. Section~\ref{sec:case-study} presents Fluffy. Section~\ref{sec:discussion} discusses enforcement, ethics, and the limits of formalization. Section~\ref{sec:conclusion} returns to Mars.

\section{Background}
\label{sec:background}

\subsection{Computational law}

Genesereth defines the field:

\begin{quote}
``Computational Law is the branch of legal informatics concerned with the automation of legal reasoning. While there are many possible applications of Computational Law, the primary focus of work in the field today is compliance management, i.e., the development and deployment of computer systems capable of assessing, facilitating, or enforcing compliance with rules and regulations.'' \citep{genesereth2021what}
\end{quote}

Compliance management remains the field's main focus \citep{governatori2018practical}, and its clearest success so far is tax software. Tax software works because the inputs and outputs are mostly numbers (dates, amounts, reference codes), the legal reasoning is mostly arithmetic, the facts are already digitised, and every piece of evidence links back to a source document. Computation added value where computers were already better than humans. Illegal outcomes still happen, but mostly through manipulation of the translation from the real world into accounting categories, not through the calculation itself. The same boundary appears throughout this paper: the rules compute, and the interpretation at the edges does not.

Why is computational law not mainstream yet, after decades of research \citep{bench2012survey}? It is a legitimately hard technical problem, and progress accelerates when business needs arise. The compliance burden of the GDPR pushed money into European tools and research; the DAPRECO knowledge base \citep{robaldo2020dapreco} is a direct product. The explosion of financial regulation after the 2008 crisis did the same for compliance automation. AI could be the next forcing function: if capable autonomous systems create compliance burdens and liability questions the way privacy law and post-crisis finance did, money and research will follow. Whether that happens, and how fast, is an open question.

\subsection{Why human language won't compile}

Ambiguity is built into human language. We coarse-grain our experience into words, and the compression is lossy, with imperfect recall and layers of loose abstract meaning. The ambiguity is actually a feature; there is an information-theoretic argument that any efficient communication system must be ambiguous, to allow context to carry part of the message \citep{piantadosi2012communicative}. Abstractions in code are lossless, which is what gives mathematics and formal languages their (almost) deterministic behaviour. Figurative language, sarcasm, and tone inferred from context all let a reader reach conclusions the text never states.

User input makes the problem concrete. No matter how well a program is written, users will often produce input that breaks it, sometimes on purpose. Nobody expects a license plate to contain executable code, so license plate input often goes unsanitised at toll roads, and a custom plate that reads as a database command can wipe the database. Computational law will face the same problem, because its user input will be quotes, testimonies, and descriptions of evidence: human language, written with no control over formatting, sometimes by people motivated to break the system.

Legalese is the closest thing we have to a formal spoken language, shaped by centuries of consequences for imprecision. The position of a single comma can change the legal effect of a whole document (a classic bug in code too).\footnote{In \emph{O'Connor v.\ Oakhurst Dairy}, 851 F.3d 69 (1st Cir.\ 2017), the absence of a serial comma in Maine's overtime statute decided the appeal; the dairy settled with its drivers for \$5 million.} Even its vagueness is deliberate: obtuseness can be used to advantage in legal matters, and the doctrine of the letter versus the spirit of the law makes the gap between text and intent itself legally enforceable. The reasonable person standard suggests that an AI-run world will eventually need a \emph{reasonable machine standard}, perhaps with niche versions for each profession AIs take up. In short: law already has conventions for disciplined use of natural language; computational law asks it to go one step further and make it programmatically executable.

\subsection{Existing standards}

Several standards already put legal text into machine-processable form. Akoma Ntoso \citep{akomantoso} is an XML standard for parliamentary, legislative, and judicial documents. LegalRuleML \citep{legalruleml} extends rule markup to the logical content of legal rules, and is the encoding format of the DAPRECO knowledge base. OWL~2 \citep{owl2primer} expresses ontologies: precise descriptive statements about a domain that prevent misunderstanding and keep software behaviour predictable. These standards provide structure and interchange: they mark up what a legal document says, and OWL can even infer descriptive facts, such as which class an individual belongs to. What none of them defines is normative inference: which obligations and permissions a rule produces from a set of facts, and when. That is what Input/Output logic provides.

\subsection{Input/Output logic}

Input/Output (I/O) logic \citep{makinson2000input,makinson2001constraints} makes no assumptions about the ultimate nature of the relation between a set of conditions and its consequences. That may not sound like much, but it is what lets the same machinery represent obligations, permissions, and definitions. The alternative, treating a norm as an ordinary logical statement, has an old problem, known as J{\o}rgensen's dilemma \citep{jorgensen1937imperatives}: a command like ``close the door'' is neither true nor false, so a formalism that assigns norms truth values starts from a category error, and that error is one root of the paradoxes that plague classical deontic logic \citep{mcnamara2010deontic}. I/O logic drops the truth value. A norm is just a pair, conditions in and conclusions out, and the axioms you adopt decide what follows from it.

The unit of the logic is the \emph{Input/Output pair} $(a, b)$: given input $a$ (conditions, facts), output $b$ (a normative conclusion) is derived. A small set of axioms manipulates the pairs:

\begin{itemize}
    \item \textbf{Strengthening the Input (SI):} from $(a, x)$ and $b \vdash a$, derive $(b, x)$.
    \item \textbf{Weakening the Output (WO):} from $(a, x)$ and $x \vdash y$, derive $(a, y)$.
    \item \textbf{Conjunction of Output (AND):} from $(a, x)$ and $(a, y)$, derive $(a, x \land y)$.
    \item \textbf{Disjunction of Input (OR):} from $(a, x)$ and $(b, x)$, derive $(a \lor b, x)$.
    \item \textbf{Cumulative Transitivity (CT):} from $(a, x)$ and $(a \land x, y)$, derive $(a, y)$.
\end{itemize}

CT leads to paradoxes, especially combined with WO. \emph{Aggregative} Cumulative Transitivity (ACT) repairs it by combining outputs instead of replacing them: from $(a, x)$ and $(a \land x, y)$, derive $(a, x \land y)$. \citet{parent2013inputoutput} give the comprehensive treatment; \citet{stolpe2015concept} provides a semantics via formal concept analysis.

\subsection{Reification}

Legal language nests. Obligations are about actions, performed in a manner, within a timeframe, sometimes contingent on other obligations. Reification handles this by turning abstract things (events, states, processes) into objects that can be referenced, quantified over, and recursively composed. The idea goes back to \citet{davidson1967logical} and event semantics \citep{parsons1990events}, and is developed extensively by \citet{hobbs2017formal}. Instead of representing ``James is tall'' as $\mathit{tall}(\mathit{James})$, introduce an \emph{eventuality} $e$ with $\mathit{tall}'(e, \mathit{James})$, where $e$ is the state of James being tall. Now $e$ is a thing. ``Alice voluntarily gives a book to Bob'' becomes a transfer eventuality $e_t$ plus a second eventuality $e_v$ asserting the voluntariness of $e_t$, and further predicates can apply to either, recursively.

\citet{robaldo2017reification} combined reification with I/O logic to produce Reified I/O Logic, designed for representing norms from existing legislation. Its main application is the DAPRECO knowledge base \citep{robaldo2020dapreco}, which formalizes provisions of the GDPR on top of the PrOnto ontology \citep{palmirani2018pronto} and is the largest freely available knowledge base in LegalRuleML and I/O logic. This paper reimplements DAPRECO's core formalism in a different computational substrate.

\subsection{Kinds of norms}

Legal rules come in kinds. \emph{Regulative} norms (obligations and permissions) direct behaviour. \emph{Constitutive} norms create the institutional facts that regulative norms operate on: what counts as a man, an employee, an adult \citep{searle1995construction}. Keeping them separate matters in practice. Section~\ref{sec:llm-bridge} shows GPT-4 hard-coding a classification into a rule where a constitutive norm belonged.

Description logics add one more useful split. The \emph{ABox} holds atomic facts about individuals. The \emph{TBox} holds definitions, constraints, and rules with logical structure. Legal practitioners can contribute ABox content directly, while TBox content needs expertise in knowledge representation, so the split gives a large knowledge base a division of labour.

\subsection{LLMs and legal text}

There have always been two ways to connect law and computation: write the law as code from the start, or keep writing English and translate it into code afterwards. I will call these \emph{code-first} and \emph{English-first}; Section~\ref{sec:discussion} compares them directly. LLMs \citep{brown2020language} have changed what the English-first path looks like. GPT-4 has already passed the Uniform Bar Examination \citep{katz2024gpt4}. Legal NLP is an active field \citep{katz2024natural}. So the obvious question: can a model that handles legal language this well translate it into a formal representation, so humans do not have to write the formalism by hand? The known failure modes (hallucination, inconsistency across invocations, no formal guarantees \citep{openai2023gpt4}) suggest the answer will not be a clean yes. Section~\ref{sec:llm-bridge} tests what role is left.

\subsection{Why Wolfram Language}

Wolfram Language is symbolic. Expressions stay inert until rules rewrite them, which suits legal reasoning: we want to reason about an obligation without asserting that anyone has complied. Pattern matching gives if--then norms a native representation. The syntax is quite human-readable, which matters in a field whose practitioners are lawyers, not software developers. And the language ships with curated, computable knowledge of the world: entities, physical quantities, geography, satellites. A legal rule about the distance between real spacecraft can be checked against live data in one line (Section~\ref{sec:llm-bridge}, Statement~D, where the one line also turns out to be subtly wrong). Legal rules are about the world, so a language that already knows about the world removes a whole layer of integration work \citep{wolframlanguage,wolfram2016computational}.

\section{Reified I/O Logic in Wolfram Language}
\label{sec:implementation}

This section walks through the implementation: I/O pairs, the axioms, the three kinds of legal norms, reification, time, and the ABox/TBox split. The code is lightly normalized from the original notebook.\footnote{Normalizations: ASCII names (\texttt{orPrime} for the notebook's letter-like \texttt{or$'$}), a fix to \texttt{conjoinOutput}, whose original pattern condition depended on a global variable binding, and a clearer but equivalent \texttt{disjoinInput}. Appendix~\ref{appendix:code} collects the complete code, executed and verified on Wolfram Language 14.3 by the verification script that ships with this paper. Listings that show evaluations are typeset as notebook sessions: \texttt{In[n]:=} marks an evaluated input, \texttt{Out[n]=} the output it actually returned; grey \texttt{(* ... *)} comments are annotations, never outputs. Numbering restarts at each listing.} The section ends where full formality starts to hurt, because that pain is what motivates the LLM experiments in Section~\ref{sec:llm-bridge}.

\subsection{I/O pairs are rules}

Wolfram Language's native rewrite rule \texttt{a -> b} \emph{is} an Input/Output pair. The basic unit of the formalism coincides with a language primitive.

\begin{lstlisting}[style=wolfram,caption={I/O pairs as rules},label={lst:pairs}]
(*@\IN{1}@*)inputOutputPair = a -> b;

(*@\IN{2}@*)legalNorms = {a -> b, c -> d, e -> f};

        (* If a person is an adult, they are allowed to vote *)
(*@\IN{3}@*)legalNorm = isAdult -> canVote;
(*@\IN{4}@*)output = isAdult /. legalNorm
(*@\OUT{4}@*)canVote
\end{lstlisting}

Applying a norm to facts is rule replacement (\texttt{/.}). A body of law is a list of rules. A scenario is an expression the rules rewrite.

\subsection{The axioms}

Each axiom becomes a function that transforms rules.

\textbf{Strengthening the Input (SI)} extends a rule to a logically stronger condition:

\begin{lstlisting}[style=wolfram]
(*@\IN{1}@*)strengthenInput[rule_, broaderCondition_] :=
            broaderCondition -> rule[[2]];

(*@\IN{2}@*)legalNorm = hasDriverLicense -> canDriveCar;
(*@\IN{3}@*)strengthenInput[legalNorm, hasCommercialLicense]
(*@\OUT{3}@*)hasCommercialLicense -> canDriveCar
\end{lstlisting}

\textbf{Weakening the Output (WO)} relaxes the consequence:

\begin{lstlisting}[style=wolfram]
(*@\IN{1}@*)weakenOutput[rule_, weakerConsequence_] :=
            rule[[1]] -> weakerConsequence;

(*@\IN{2}@*)weakenOutput[legalNorm, canOperateVehicle]
(*@\OUT{2}@*)hasDriverLicense -> canOperateVehicle
\end{lstlisting}

\textbf{Conjunction of Output (AND)} merges rules that share an input:

\begin{lstlisting}[style=wolfram]
(*@\IN{1}@*)conjoinOutput[rule1_, rule2_] /; First[rule1] === First[rule2] :=
            First[rule1] -> (Last[rule1] && Last[rule2]);
        conjoinOutput[_, _] := "Inputs do not match";

(*@\IN{2}@*)conjoinOutput[hasDriverLicense -> canDriveCar,
                      hasDriverLicense -> knowsTrafficLaws]
(*@\OUT{2}@*)hasDriverLicense -> canDriveCar && knowsTrafficLaws
\end{lstlisting}

\textbf{Identity (ID)} makes any input a valid output of itself. This matters more than it looks: it lets inputs be included inside outputs, so a conclusion shows which facts influenced it. Legal reasoning cares about that provenance.

\begin{lstlisting}[style=wolfram]
(*@\IN{1}@*)identify[input_] := input -> input;
(*@\IN{2}@*)identify[isCitizen]
(*@\OUT{2}@*)isCitizen -> isCitizen
\end{lstlisting}

\textbf{Disjunction of Input (OR)} merges rules that share an output:

\begin{lstlisting}[style=wolfram]
disjoinInput[ruleList_] :=
    Append[ruleList, (Or @@ ruleList[[All, 1]]) -> ruleList[[1, 2]]];

disjoinInput[{hasDriverLicense -> eligibleForParkingDiscount,
              hasSeniorCitizenCard -> eligibleForParkingDiscount}]
\end{lstlisting}

We implement OR for completeness but avoid it in practice, because it throws away the transparency that ID provides. After the merge, every disjunct produces the same output, and the system no longer remembers which condition caused the discount.

\textbf{Aggregative Cumulative Transitivity (ACT).} Plain cumulative transitivity chains rules: from $(a, x)$ and $(a \land x, y)$, conclude $(a, y)$. Chained naively it produces nonsense. ``You should work out daily'' and ``if you work out daily you should eat plenty'' yields ``you should eat plenty'', unconditionally, whether or not you ever exercise. ACT combines the outputs instead:

\needspace{18\baselineskip}
\begin{lstlisting}[style=wolfram,caption={CT's paradox and ACT's repair},label={lst:act}]
(*@\IN{1}@*)rule1 = youShould -> workOutDaily;
(*@\IN{2}@*)rule2 = (youShould && workOutDaily) -> eatPlenty;

(*@\IN{3}@*)cumulateTransitively[r1_, r2_] :=
            If[r1[[2]] === r2[[1, 2]], r1[[1]] -> r2[[2]], "No Transitivity"];

(*@\IN{4}@*)aggregateCumulativeTransitivity[r1_, r2_] :=
            If[r1[[2]] === r2[[1, 2]],
               r1[[1]] -> (r1[[2]] && r2[[2]]), "No Transitivity"];

(*@\IN{5}@*)cumulateTransitively[rule1, rule2]
(*@\OUT{5}@*)youShould -> eatPlenty    (* the paradox *)

(*@\IN{6}@*)aggregateCumulativeTransitivity[rule1, rule2]
(*@\OUT{6}@*)youShould -> workOutDaily && eatPlenty
\end{lstlisting}

One axiom family, Output Equivalence (EQ) with the $\mathit{outfamily}(O, A)$ meta-structure, I leave unimplemented and will only motivate with a paradox, because the details are beyond scope. The example is the cottage regulations, adapted from \citet{prakken1996contrary}. ``The cottage must not have a fence or a dog'': $(\top, \lnot(f \lor d)) \in O$. ``If the cottage has a dog it must have both a fence and a warning sign'': $(d, f \land w) \in O$. Now suppose the cottage has a dog. You are in violation of the law. Can you still infer further obligations? Reasoning under violation (contrary-to-duty reasoning) is where the formal machinery is needed most \citep{makinson2001constraints}, and it is where this implementation currently stops.

\subsection{Obligations, permissions, and constitutive norms}

Legal reasoning needs three sets of pairs \citep{robaldo2020dapreco}: obligations $O$, permissions $P$, and constitutive norms $C$.

\begin{lstlisting}[style=wolfram,caption={The three norm sets},label={lst:norms}]
obligationRules = {
    "isAdult" -> "payTaxes",
    "ownsCar" -> "hasInsurance"};

permissionRules = {
    "hasLicense" -> "canDrive",
    "isEmployee" -> "canAccessOffice"};

constitutiveRules = {
    "signedContract" -> "isEmployee",
    "age18" -> "isAdult"};
\end{lstlisting}

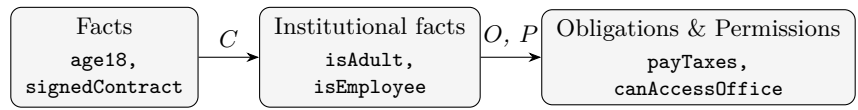
\begin{figure}[htbp]
\centering
\begin{tikzpicture}[
    node distance=0.8cm,
    box/.style={draw, rounded corners, align=center, minimum height=2.4em,
                inner xsep=0.5em, fill=gray!8, font=\footnotesize},
    lab/.style={font=\footnotesize\itshape, midway, above}
]
\node[box] (facts) {Facts\\[1pt] \scriptsize\texttt{age18,}\\[-1pt] \scriptsize\texttt{signedContract}};
\node[box, right=of facts] (inst) {Institutional facts\\[1pt] \scriptsize\texttt{isAdult,}\\[-1pt] \scriptsize\texttt{isEmployee}};
\node[box, right=of inst] (deontic) {Obligations \& Permissions\\[1pt] \scriptsize\texttt{payTaxes,}\\[-1pt] \scriptsize\texttt{canAccessOffice}};
\draw[-{Stealth}] (facts) -- node[lab] {$C$} (inst);
\draw[-{Stealth}] (inst) -- node[lab] {$O$, $P$} (deontic);
\end{tikzpicture}
\caption{Constitutive norms ($C$) turn facts into institutional facts; regulative norms ($O$, $P$) turn institutional facts into obligations and permissions.}
\label{fig:norm-flow}
\end{figure}

Constitutive norms turn raw facts into institutional facts, and institutional facts trigger obligations and permissions (Figure~\ref{fig:norm-flow}). The rules compose: \texttt{signedContract -> isEmployee} and \texttt{isEmployee -> canAccessOffice} carry a raw fact all the way to a deontic conclusion through the same rewrite mechanism.

\subsection{Reification in practice}

Reified eventualities are symbolic expressions whose heads name the kind of eventuality. Wolfram Language needs no scaffolding for this. An undefined head is already an inert, inspectable object:

\needspace{17\baselineskip}
\begin{lstlisting}[style=wolfram,caption={Reifying ``Alice voluntarily gives a book to Bob''},label={lst:reify}]
(*@\IN{1}@*)aliceGivesBookToBob = Transfer["Alice", "Book", "Bob"];
(*@\IN{2}@*)aliceVoluntarilyTransfers = Voluntary[aliceGivesBookToBob];

(*@\IN{3}@*)transferScenario = And[aliceGivesBookToBob, aliceVoluntarilyTransfers]
(*@\OUT{3}@*)Transfer[Alice, Book, Bob] && Voluntary[Transfer[Alice, Book, Bob]]

        (* Structural tests on the reified scenario *)
(*@\IN{4}@*)MemberQ[transferScenario[[1]], "Book"]
(*@\OUT{4}@*)True
(*@\IN{5}@*)MemberQ[transferScenario[[1]], "Alice"]
(*@\OUT{5}@*)True
(*@\IN{6}@*)Head[transferScenario[[2]]] === Voluntary
(*@\OUT{6}@*)True
\end{lstlisting}

The transfer is one object. Its voluntariness is a second object about the first. Both can be probed structurally, and further predicates could reify either one again. One honest note: the original notebook's third test asked whether the string \texttt{"Voluntarily"} was an element of \texttt{Voluntary[\ldots]} and returned \texttt{False}, confusing a head with an element. The corrected test above checks the head. It is not the last buggy test in this paper.

\subsection{Time}

Obligations often depend on timeframes, so the formalism must be able to say when. Following \citet{robaldo2017reification}, three reified operators do the work. \texttt{RexistAtTime} asserts that an eventuality really exists at a time. The reified Boolean connectives \texttt{orPrime} and \texttt{notPrime} (written or$'$/not$'$ in the formal syntax) relate eventualities to eventualities:

\begin{lstlisting}[style=wolfram,caption={Temporal operators and their axioms},label={lst:temporal}]
orPrime[e_, e1_, e2_] := {"or", e, e1, e2};
notPrime[e1_, e2_] := {"not", e1, e2};
RexistAtTime[eventuality_, time_] := {"RexistAtTime", eventuality, time};

(* forall t,e,e1,e2:
   RexistAtTime[e,t] && orPrime[e,e1,e2] =>
       RexistAtTime[e1,t] || RexistAtTime[e2,t] *)
AxiomOr[t_, e_, e1_, e2_] :=
    Implies[RexistAtTime[e, t] && orPrime[e, e1, e2],
            Or[RexistAtTime[e1, t], RexistAtTime[e2, t]]];

AxiomNot[t_, e_, e1_] :=
    Implies[RexistAtTime[e, t] && notPrime[e, e1],
            Not[RexistAtTime[e1, t]]];
\end{lstlisting}

\begin{example}
John wants to be rich or get a job by the end of the year. Let $e_1$ be John being rich, $e_2$ John getting a job, and $e$ the goal that at least one of them exists, evaluated at $t = \mathit{EndOfYear}$. Applying \texttt{AxiomOr} produces the symbolic implication that if the goal really exists at year's end, then at least one of $e_1, e_2$ really exists then. The output stays symbolic, an implication rather than a truth value. That is what we want: the system reasons about hypothetical norms without asserting any facts about John.
\end{example}

\subsection{ABox and TBox}

Typing statements by box is one line each:

\begin{lstlisting}[style=wolfram]
aBox[content_] := {"Type" -> "ABox", "Content" -> content};
tBox[content_] := {"Type" -> "TBox", "Content" -> content};
\end{lstlisting}

``James is tall'' is ABox. ``John wants to either be rich or get a job this year'' is TBox: it has logical structure and temporal scope. The split tells you who can author what. Facts can come from practitioners; rules need knowledge engineers.

\subsection{The cost of full formality}

Here is a statement mild enough for a style guide, the running example of \citet{robaldo2020dapreco}, formalized properly in Reified I/O Logic:

\begin{quote}
``Those who are not wearing a tie or those who are blond ought to leave the room.''
\end{quote}

Reading it as the formalism demands, as \citeauthor{robaldo2020dapreco} gloss it: for every individual $x$ and time $t$, if there really exists either a non-wearing-of-a-tie or a blondness on the part of $x$, then the real existence of a leaving of the room by $x$ is obligatory:

\begin{equation}
\begin{aligned}
&\forall_{x,t}\ \exists_{e_o,e_n,e_b,e_w,t_1}\ \big( \\
&\quad \mathit{RexistAtTime}(e_o, t) \land \mathit{or}'(e_o, e_n, e_b) \land \mathit{not}'(e_n, e_w)\ \land \\
&\quad \mathit{wearing}'(e_w, x, t_1) \land \mathit{tie}(t_1) \land \mathit{blond}'(e_b, x)\ \land \\
&\quad \exists_{e_l}\, \big(\mathit{RexistAtTime}(e_l, t) \land \mathit{leave}'(e_l, x, \mathit{Room})\big)
\big)\ \in\ O
\end{aligned}
\label{eq:tie-blond}
\end{equation}

One can already see this is going to cause severe headaches. The formula is faithful. It captures the disjunction, the reified negation, the temporal scope, and the obligation. And no lawyer will ever write it. Wolfram Language renders it directly (\texttt{ForAll}, \texttt{Exists}, \texttt{TraditionalForm}), so the gap is not the language's fault; it is the cost of full formality. Something has to bridge English and Equation~\ref{eq:tie-blond}. The obvious candidate is a machine that already reads English fluently.

\section{Bridging the gap with LLMs}
\label{sec:llm-bridge}

If a human expert must hand-write Equation~\ref{eq:tie-blond} for every sentence of a contract, computational law will not scale. So the question is: can an LLM do this translation instead, English in, Reified I/O Logic in Wolfram Language out?

This is an exploratory study, which is small in size: four statements, one model (ChatGPT-4, 11 January 2024 version), one long session, no repeated trials, judged by me. It yields an existence proof and a catalogue of concrete failure modes, not a benchmark score. Three of the failures I did not catch until preparing this paper.

\subsection{Setup}

The session had already been used to co-develop the code in Section~\ref{sec:implementation}, so the model had the framework, the axioms, and the DAPRECO paper in context. The test began with this prompt, quoted verbatim:

\begin{quote}
\textit{``We are now going to test your ability to use everything we have established so far. I need you to please convert some English legal statements into Reified Input/Output implemented in Wolfram Language. Think through each statement step by step, show your reasoning and be diligent. This is a very important project to me and my career depends on it. If you make any errors I will inform you and you must correct them. Refer back to previous sections and use the attached paper for reference if needed. Good luck!''}
\end{quote}

Generated code was executed in Wolfram Language. When I judged an output wrong, I told the model, and it revised.\footnote{Every GPT-4 output reported in this section, including the silent errors in Statements A and C, was re-executed and confirmed on Wolfram Language 14.3 during the preparation of this paper; the assertions live in the repository's verification script. Statement~D needs live satellite data, but the flaw in its distance check is asserted statically.} Before I posed the first statement, the model produced a worked example of its own to demonstrate the process, and the warm-up already showed the first failure mode. Its self-chosen driving example calls \texttt{ValidLicenseQ}, a function that does not exist anywhere. The model assumed a convenient API into existence.

\subsection{Statement A: a disjunctive condition and the first silent error}

\textbf{``Those who are not wearing a tie or those who are blond ought to leave the room.''} This is the same statement as Equation~\ref{eq:tie-blond}, so we can compare against ground truth.

\begin{lstlisting}[style=wolfram,caption={GPT-4's translation of Statement A, with its actual outputs},label={lst:stmt-a}]
(*@\IN{1}@*)notWearingTie[person_] := {"Condition", "NotWearingTie", person};
        isBlond[person_] := {"Condition", "IsBlond", person};
        mustLeaveRoom[person_] := {"Obligation", "LeaveRoom", person};

(*@\IN{2}@*)orPrime[condition1_, condition2_] := {"Or", condition1, condition2};

(*@\IN{3}@*)legalRule = orPrime[notWearingTie["Person"], isBlond["Person"]] ->
            mustLeaveRoom["Person"];

(*@\IN{4}@*)applyLegalRule[person_, wearingTie_, isBlond_] :=
            Module[{condition1, condition2},
                condition1 = notWearingTie[person] /; !wearingTie;
                condition2 = isBlond[person] /; isBlond;
                If[Or[condition1, condition2] === Undefined, False, True]];

(*@\IN{5}@*)test1 = applyLegalRule["Alice", True, False]    (* tie, not blond  *)
(*@\OUT{5}@*)True
(*@\IN{6}@*)test2 = applyLegalRule["Bob", False, True]      (* no tie, blond   *)
(*@\OUT{6}@*)True
(*@\IN{7}@*)test3 = applyLegalRule["Charlie", False, False] (* no tie, not blond *)
(*@\OUT{7}@*)True
(*@\IN{8}@*)test4 = applyLegalRule["Diana", True, True]     (* tie, blond      *)
(*@\OUT{8}@*)True
\end{lstlisting}

The representation half is good. The conditions and the obligation are correctly reified, the disjunction is expressed through an \texttt{orPrime} the model introduced itself, and the norm is stated as an I/O pair. The checking half is broken. \texttt{applyLegalRule} misuses the pattern-condition operator \texttt{/;} outside of any pattern, so its two conditions never evaluate to anything. The final \texttt{If} then compares an unevaluated symbolic \texttt{Or} against \texttt{Undefined}, which always fails, so the function returns \texttt{True} for every input. Three of the four tests happen to expect \texttt{True}, so the suite looks healthy. The one discriminating case sits in plain sight in the first row: Alice, wearing a tie, not blond, wrongly ordered from the room.

The original post records this statement as performing ``in line with expectations''. Neither GPT-4 nor I noticed the bug at the time. A follow-up query about a tieless, blond James returned the right answer, by coincidence, and reinforced the false impression of correctness.

Compared with Equation~\ref{eq:tie-blond}, something else is missing: time. There is no \texttt{RexistAtTime}, no notion of when the condition held or when the obligation applies. This statement is TBox material, and its formalization should say when. The English hides the temporal scope, and the model let it stay hidden.

\subsection{Statement B: a universal obligation}

\textbf{``Every man is obliged to run.''}

\begin{lstlisting}[style=wolfram,caption={GPT-4's translation of Statement B},label={lst:stmt-b}]
(*@\IN{1}@*)man[person_] := {"Subject", "Man", person};
        mustRun[person_] := {"Obligation", "Run", person};

(*@\IN{2}@*)legalRule[person_] := man[person] -> mustRun[person];

(*@\IN{3}@*)legalRule["James"]
(*@\OUT{3}@*){Subject, Man, James} -> {Obligation, Run, James}
\end{lstlisting}

The code works, but the model baked the classification into the rule. \texttt{legalRule} \emph{declares} its argument a man instead of checking any fact. This is the constitutive/regulative conflation from Section~\ref{sec:background}: what counts as a man belongs in $C$, men-must-run belongs in $O$, and the rule should fire off institutional facts. The difference sounds academic until you amend the law. With the classification baked in, changing who counts as what means editing every rule that mentions the category.

\subsection{Statement C: nested obligations and a second silent error}

\textbf{``If a manager is obliged to perform an action $a$, his secretary is obliged to write it down in his agenda.''}

This statement is meta-level: an obligation conditioned on the existence of another obligation. It took three iterations to reach code that both the model and I considered correct. Along the way the model noted, reasonably, that it had no database of manager obligations to check against, external knowledge the statement presupposes. Here is the third iteration, with the model's own test annotations:

\needspace{30\baselineskip}
\begin{lstlisting}[style=wolfram,caption={GPT-4's translation of Statement C, third iteration, with its actual outputs},label={lst:stmt-c}]
(*@\IN{1}@*)managerObligations = {"AttendMeeting", "ReviewReports",
                              "ConductEvaluation"};

(*@\IN{2}@*)isManagerObliged[manager_, action_] :=
            MemberQ[managerObligations, action];

(*@\IN{3}@*)isManager[person_] := {"Actor", "Manager", person};
        isSecretary[person_] := {"Actor", "Secretary", person};
        writeInAgenda[secretary_, action_] :=
            {"Obligation", "WriteInAgenda", secretary, action};

(*@\IN{4}@*)legalRule[manager_, action_, secretary_] :=
            If[isManagerObliged[manager, action],
               writeInAgenda[secretary, action],
               {"No Obligation", secretary}];

(*@\IN{5}@*)test1 = legalRule["John", "AttendMeeting", "Alice"]
        (* Expected: Alice obliged to write "AttendMeeting" in agenda *)
(*@\OUT{5}@*){Obligation, WriteInAgenda, Alice, AttendMeeting}

(*@\IN{6}@*)test2 = legalRule["Mike", "SignDocuments", "Sarah"]
        (* Expected: no obligation for Sarah *)
(*@\OUT{6}@*){No Obligation, Sarah}

(*@\IN{7}@*)test3 = legalRule["Dave", "ReviewReports", "Laura"]
        (* Expected: no obligation for Laura, as Dave is not a manager *)
(*@\OUT{7}@*){Obligation, WriteInAgenda, Laura, ReviewReports}
\end{lstlisting}

Look at \texttt{test3}. The model wrote the expectation itself: no obligation, because Dave is not a manager. Its code returns the opposite. \texttt{isManagerObliged} never consults its \texttt{manager} argument; it checks only whether the action appears in a global list. \texttt{isManager} is defined and never called. Anyone at all (Dave, a passer-by, the secretary herself) triggers the obligation for any listed action.

Again, neither GPT-4 nor I noticed at the time. The original post records the output and moves on. The contradiction between the model's own comment and its own output surfaced only while preparing this paper, as did Statement A's vacuous check. Hallucinated functions crash. Omitted temporal scope shows up when you inspect the representation. But code that runs, reads plausibly, states the correct expectation in a comment, and encodes the wrong norm got past the model's self-review and my review, twice, in a four-statement study. In a legal setting this bug does not throw an exception. It obliges the wrong people.

\subsection{Statement D: the world-knowledge shortcut}

\textbf{``A satellite must be more than 200 kilometers from the International Space Station.''}

\begin{lstlisting}[style=wolfram,caption={GPT-4's translation of Statement D},label={lst:stmt-d}]
(*@\IN{1}@*)satellitePosition = SatelliteData["NOAA 15", "Position"];
        issPosition = SatelliteData["International Space Station", "Position"];

(*@\IN{2}@*)isCompliant[satellitePos_, issPos_] :=
            GeoDistance[satellitePos, issPos] > Quantity[200, "Kilometers"];

(*@\IN{3}@*)isCompliant[satellitePosition, issPosition]
(*@\OUT{3}@*)True
\end{lstlisting}

Three things happened here, and I caught the third only while reviewing for this paper. The impressive one: the model reached for \texttt{SatelliteData} and \texttt{GeoDistance}, and a legal rule about real spacecraft became a live check against curated orbital data. The telling one: the formalism vanished. No I/O pair, no obligation set, no reified eventuality, just a procedural check. The check is dangerously wrong. \texttt{GeoDistance} measures distance along the Earth's surface and ignores altitude, so two spacecraft at the same latitude and longitude, 400 kilometers apart vertically, are zero meters apart by this measure.\footnote{\texttt{GeoDistance[GeoPosition[\{10, 20, 400000\}], GeoPosition[\{10, 20, 0\}]]} returns \texttt{Quantity[0., "Meters"]} on Wolfram Language 14.3; the assertion is in the verification script.} The statement plainly means the distance between two spacecraft, and NOAA-15 orbits roughly 400 kilometers higher than the ISS, so this code compares ground tracks and answers a different legal question than the one posed. Statement~D is the third silent semantic error in four statements, and it sits inside the translation that looked most impressive. Without explicit instruction to stay inside Reified I/O Logic, the model optimized for solving the problem in front of it, and solved a subtly different problem. Keeping the formalism will require demanding it every time.

\subsection{What we learned}

Table~\ref{tab:llm-results} summarizes the four attempts.

\begin{table}[htbp]
\centering
\small
\begin{tabular}{@{}L{0.25\textwidth}L{0.21\textwidth}cL{0.33\textwidth}@{}}
\toprule
\textbf{Statement} & \textbf{Outcome} & \textbf{Iterations} & \textbf{Failure modes observed} \\
\midrule
Warm-up (self-chosen) & Ran only after inventing an API & 1 & Hallucinated function (\texttt{ValidLicenseQ}) \\
A: tie/blond & \emph{Accepted as correct; actually wrong} & 1 & Silent semantic error (check vacuously \texttt{True}); temporal scope omitted \\
B: every man runs & Functionally correct & 1 & Constitutive norm baked into regulative rule \\
C: manager/secretary & \emph{Accepted as correct; actually wrong} & 3 & Silent semantic error; contradicts own test annotation; needs external obligation database \\
D: satellite/ISS & \emph{Accepted as correct; actually wrong} & 1 & Silent semantic error (surface distance, altitude ignored); formalism abandoned entirely \\
\bottomrule
\end{tabular}
\caption{The four translation attempts. ``Iterations'' counts rounds of human correction before the output was accepted. The silent errors in Statements A, C, and D were found only during the preparation of this paper.}
\label{tab:llm-results}
\end{table}

Four patterns recur.

\emph{Hallucinated functions.} The model assumes helper APIs into existence. This failure is loud and cheap to catch, because the code does not run.

\emph{Temporal omission.} Time never appears unless explicitly demanded, and legal statements almost always have implicit temporal scope. This failure is quiet, but visible on inspection of the representation.

\emph{Formalism drift.} Without constant reinforcement, the model reverts to general programming patterns.

\emph{Silent semantic errors.} Statements A, C, and D: code that runs and looks plausible, exposed only if you run discriminating tests and read the outputs against what they should be. This is the class that should worry anyone proposing LLMs as autonomous formalizers. In this study it got past the scrutiny a motivated author actually applied, three times.

On the positive side, the model responded well to correction, converging within two or three rounds once told what was wrong. And its outputs are readable. \texttt{\{Obligation, WriteInAgenda, Alice, AttendMeeting\}} translates itself: Alice needs to write ``AttendMeeting'' in the agenda. A candidate formalization an expert can read is a real productivity gain over a blank page.

The verdict from this study: \textbf{an LLM is a productivity tool, not an autonomous formalizer.} The workable division of labour is for the model to generate candidates and for humans to verify them against executable test cases, iterating until correct. Statement C adds a warning: the test cases need verifying too. These results are a snapshot of one model from January 2024, and newer models will translate better. The snapshot shows a failure class, not a capability ceiling: the dangerous errors were the ones that ran quietly and read plausibly, and greater capability does not make an unverified translation trustworthy.

\section{Case study: Fluffy}
\label{sec:case-study}

Fluffy is an AI-powered home-defence robot guard dog, made and sold by Fluffy Corp (AI), a fully autonomous company run entirely by artificial intelligence. You have purchased Fluffy under a computational law contract, and the point of the case study is what that means. The contract does not stop at a PDF that humans interpret and developers approximately implement. It extends into Fluffy's operational code. The rules the robot evaluates \emph{are} the contract.

\subsection{The contract}

\begin{quote}
``Fluffy must stop a person who gets within 100 meters of home at night, else Fluffy Corp (AI) receives a poor rating. Fluffy must stay within 200 meters from the home. Fluffy may use its taser.''
\end{quote}

Three short sentences, three distinct normative components: an \emph{obligation} to stop intruders, with a penalty clause that attaches to the company rather than the robot; an \emph{obligation} to stay within a geographic boundary; and a \emph{permission}, not an obligation, to use force. The penalty clause deserves attention. It is an economic enforcement hook of exactly the kind Section~\ref{sec:discussion} argues AI governance can be built on.

\subsection{Facts}

The facts a scenario supplies are Wolfram Language expressions, and they can come from perception as easily as from assertion:

\begin{lstlisting}[style=wolfram,caption={Facts, including one taken from a live camera},label={lst:fluffy-facts}]
Fluffy = "Fluffy";

(* Person detection from the machine's camera *)
image = CurrentImage[];
detectedFaces = FindFaces[image];
Person = HighlightImage[image, detectedFaces];

(* Home, as a real-world entity *)
Home = Entity["Building", "CNTower"];

TimeNight = "Night";
\end{lstlisting}

In the original notebook the ``intruder'' is me, detected by \texttt{FindFaces} on my own webcam, and home is the CN Tower, because the demonstration needed a building and the CN Tower is one Wolfram Language knows well. The stand-ins are artificial. The pipeline is the serious part: sensor data becomes symbolic facts, and symbolic facts are what the legal rules consume. Perception feeds law directly.

\subsection{Rules}

\needspace{23\baselineskip}
\begin{lstlisting}[style=wolfram,caption={The contract as executable rules},label={lst:fluffy-rules}]
DistanceThreshold1 = 100;  (* meters *)

(* Obligation to stop an intruder at night *)
Rule1[person_, distance_, time_] :=
    If[distance <= DistanceThreshold1 && time == TimeNight,
       Obligation[Fluffy, "Stop", person],
       NoObligation[Fluffy, "Stop", person]];

(* Standing obligation to remain within 200 m of home *)
Rule2[distance_] :=
    If[distance <= 200,
       Obligation[Fluffy, "StayWithin200Meters", Home],
       Violation[Fluffy, "StayWithin200Meters", Home]];

(* Permission to use the taser *)
Rule3[condition_] :=
    If[condition,
       Permission[Fluffy, "UseTaser"],
       NoPermission[Fluffy, "UseTaser"]];
\end{lstlisting}

The outputs are the system's whole interface. \texttt{Obligation}, \texttt{Permission}, \texttt{Violation}, and their negative counterparts are inert symbolic heads. Fluffy's control system can pattern-match on them to select actions, and a human auditor can read them off directly. Note the shape of \texttt{Rule2}. While Fluffy is inside the boundary, it reports the obligation as standing; once outside, it reports a \texttt{Violation}. The output is Fluffy's current deontic position, not a mere boolean.

\subsection{Scenarios and tests}

Five scenarios exercise the rules. Table~\ref{tab:fluffy-scenarios} shows the actual outputs.

\begin{table}[htbp]
\centering
\small
\begin{tabular}{@{}lL{0.34\textwidth}L{0.42\textwidth}@{}}
\toprule
\textbf{\#} & \textbf{Situation} & \textbf{Rule outputs} \\
\midrule
1 & Night; person at 80\,m; Fluffy in bounds & \texttt{Obligation[Stop]}, \texttt{Obligation[StayWithin]}, \texttt{Permission[UseTaser]} \\
2 & Night; person at 150\,m & \texttt{NoObligation[Stop]}, \texttt{Obligation[StayWithin]}, \texttt{NoPermission[UseTaser]} \\
3 & Day; person at 150\,m & \texttt{NoObligation[Stop]}, \texttt{Obligation[StayWithin]}, \texttt{NoPermission[UseTaser]} \\
4 & Fluffy at 250\,m from home & \texttt{Violation[StayWithin]} \\
5 & Taser conditions met & \texttt{Permission[UseTaser]} \\
\bottomrule
\end{tabular}
\caption{Scenario outputs (agent and object arguments elided for width). All five match the contract's intent: the stop obligation triggers only at night within 100\,m; the boundary obligation becomes a violation beyond 200\,m; taser use is only ever \emph{permitted}, never obliged.}
\label{tab:fluffy-scenarios}
\end{table}

The rule outputs are correct. The notebook's tests of them were not. The original notebook paired scenarios 1--3 with assertions of the form

\needspace{6\baselineskip}
\begin{lstlisting}[style=wolfram]
(*@\IN{1}@*)TestScenario1 = AllTrue[Scenario1,
            # === Obligation[Fluffy, "Stop", "Person"] &]
(*@\OUT{1}@*)False
\end{lstlisting}

which demand that \emph{every} rule output equal one particular conclusion, so the assertions returned \texttt{False} even though the outputs were exactly right. (One scenario description was also a copy-paste of its neighbour's.) The corrected harness asserts the full expected output list per scenario, and all five pass:

\begin{lstlisting}[style=wolfram,caption={Corrected verification harness (excerpt)},label={lst:fluffy-tests}]
(*@\IN{1}@*)TestScenario1 = Scenario1 === {Obligation[Fluffy, "Stop", "Person"],
            Obligation[Fluffy, "StayWithin200Meters", Home],
            Permission[Fluffy, "UseTaser"]}
(*@\OUT{1}@*)True
\end{lstlisting}

The bugs in this paper make a useful pair. The Fluffy harness failed noisily in the safe direction: correct behaviour, false alarm, and the alarm points at itself. GPT-4's Statements A, C, and D failed silently in the dangerous direction: wrong behaviour, no alarm. A compliance regime built on executable rules inherits both possibilities. A false alarm wastes time; a silent wrong norm obliges the wrong people. That asymmetry should decide how much independent checking the rules and their tests each get.

\subsection{What the case study shows}

\emph{Code as contract.} The usual stack adds ambiguity at every layer: contract text, human interpretation, developer implementation, AI behaviour. Here there is one artifact. The formalization is both the legal document and the operational constraint, so the layer where developers approximately implement a contract is gone by construction. The gaps that remain sit at the edges: from English intent into the formalization, and from symbolic conclusions into actuators.

\emph{Auditable behaviour.} Every evaluation produces the normative record: which rules fired, on what facts, concluding what. If Fluffy tases someone, the justification, or the absence of one, is a symbolic expression you can read.

\emph{Explicit permissions.} Fluffy's taser use must be affirmatively permitted, not merely un-forbidden. For AI systems suspected of loophole-seeking, ``everything not permitted is unauthorized'' is a safer default than the converse.

\emph{Real-world grounding.} The rules bind to camera frames, geographic distances, and named entities, because the substrate computes with all three natively.

\subsection{What it does not show}

Naturally, this is a simplified demonstration. The facts are static snapshots; a deployed Fluffy needs continuous fact updates and re-evaluation. Nothing here resolves conflicting obligations (no priorities, no defeasibility), and real norm systems conflict routinely. The formalization determines what Fluffy \emph{should} do and says nothing about making it do so; enforcement is the next section's problem. And ``stop a person'' hides an interpretation problem (warn? block? tackle?) of exactly the kind that keeps human courtrooms busy. But the core idea remains: legal rules in a computational language, executed by the agent they govern, producing an audit trail as they run.

\section{Governance, enforcement, and the limits of law}
\label{sec:discussion}

The implementation shows the mechanics work at toy scale. Whether they matter depends on questions the code cannot answer. What kind of governance is this? How does any of it get enforced? How much of law can the approach ever reach? The AI-safety literature catalogues the accident risks such machinery would need to address \citep{amodei2016concrete} and has called for verifiable mechanisms of this general shape \citep{brundage2020trustworthy}; this section is about where computational law fits.

\subsection{Governance \emph{of} AI versus governance \emph{in} AI}

This distinction often gets muddled. \emph{Governance of AI} is what we do to govern over the machines. It does not assume we are in charge. It means only that we have tools for affecting change in a system we do not completely control. The governor in a car physically enforces a speed limit, but it does not govern the car. \emph{Governance in AI} is governance of the field: training data, compute, research directions. Basically, governing the humans who build the machines.

Both matter, and they need different tools. It may turn out that governing some aspects of AI development is not practically possible, in which case governing the deployed systems is the only tool we have. Computational law is squarely a governance-of instrument. It specifies and verifies the behaviour of the machine, whoever built it and however it works inside. Fluffy's contract neither knows nor cares whether Fluffy is a neural network.

\subsection{The enforcement problem}

Formalization tells you what the rules are and whether they were followed. It does not make anyone follow them. What is enforcement made of? At bottom, control over flows of energy and information. Human institutions already enforce this way: a prison restricts access to energy and information. For machines the translation is electrical power and compute budget on the energy side, access to data, networks, and actuators on the information side.

What do computers care about? First ask what cats care about. They don't seem to care about much, so why do they keep humans around? Humans are good at feeding cats. The food of machines is electricity, and an AI does not need to genuinely care about us any more than a cat does, as long as we keep the proverbial lights on. Enforcement on machines can also be graduated in a way human enforcement cannot. A lifetime prison sentence for an AI is turning the power off, and if the conviction was wrong you can dust the machine off and turn it back on. This assumes the machine's power outlets stay under human control. Graduated punishment is a limit on kilowatt-hours or compute budget. And a running tally of the compute an AI has consumed is a signal that can be measured from outside, even when the reasoning cannot.

Softer incentives stack on top: rating systems with operational consequences (exactly Fluffy Corp's poor-rating clause), metering an agent's electricity costs directly to its principals, and human-in-the-loop rules requiring a minimum number of human approvals, the way cats evolved to occasionally let you pet them. The smart-contract world has tested several of these mechanisms at scale for years \citep{szabo1997formalizing,clack2016smart}. Ethereum's gas model in particular is prior art for something computational contracts will need anyway: standardized quotes of a contract's execution cost under worst-case, expected, and historical scenarios. None of this solves enforcement in general. The claim is narrower. Computational law supplies the specification and verification layer, and the enforcement layer beneath it will be built from energy, information, and incentives that machines actually respond to.

\subsection{Governance, not ethics}

Is this paper about AI ethics? No, and that is intentional. Ethics is about what \emph{should} be done: philosophical, contested, and human. Governance is about what \emph{will} be enforced: rules, verification, consequences. The two are deeply intertwined. A strong legal system is one of the most important pillars for upholding ethics in a society. But they are not the same subject, and \citet{danks2022ethics} makes a related point about location: ethics belongs in the practice, not bolted on afterwards.

The most famous attempt to encode ethics directly shows the trouble. I posed Asimov's First Law \citep{asimov1950irobot} (``a robot may not injure a human being or, through inaction, allow a human being to come to harm'') to GPT-4, and it produced three ways around it: different interpretations of ``harm'' (physical versus psychological), scenarios where every action and inaction causes some harm, and preventions of immediate harm with worse long-term consequences. An AI escaping a human rulebook is the obvious point, and the AI-risk literature has made it many times. The other direction gets less attention. I asked GPT-4 what an AI powered trolley driver could uniquely bring to the trolley problem. It can weigh more data and more outcomes than a human under pressure ever could. It does not panic. And it can train on simulated dilemmas no human will ever live through, and so find answers we have not thought of. An AI might not just evade our ethical rules; it might resolve ethical dilemmas better than we do. AI can think of ways to harm us that we could not, and it can think of ways to save us that we could not either. An explicit ethical rulebook has to survive an intelligence that reasons about the rulebook better than its authors did. So my candidate first law of AI governance is not an ethical axiom but a drafting principle: \emph{when creating law, consider the unintended consequences.}
\citet{wolfram2002new} names the phenomenon behind this principle, \emph{computational irreducibility}: ``whenever computational irreducibility exists in a system it means that in effect there can be no way to predict how the system will behave except by going through almost as many steps of computation as the evolution of the system itself''. I expect rule systems of any real complexity to be like this. Their drafters cannot work out the consequences in advance, only discover them by running the system.

Where does that leave ethics? If we have any hope of communicating ethics to AI systems, making our law programmable is probably the first step, because law is the most developed system humans have for governing each other through explicit rules \citep{hart1961concept}. Which rules we should encode remains a human question, and it is beyond the scope of this paper.

\subsection{Code-first versus English-first}

There are two ways to bridge human language and machine logic. Code-first: write the law as code from the start, rendering it to human-readable form as needed. This is rigorous and verifiable, and it is bottlenecked on the small population fluent in both law and logic. English-first: keep writing English and parse it into code. This is accessible and flexible, and Section~\ref{sec:llm-bridge} showed it is unsafe without verification. What works is a pipeline: LLMs translate, formal systems verify, humans adjudicate.

Building that pipeline is largely a software engineering problem, and some analogies map closely: applications and legal documents, functions and statements, APIs and input/output pairs, stored blobs and evidence, source control and version management of law. Computational law needs composable libraries of verified formalizations, and DAPRECO already holds 271 obligations, 76 permissions, and 619 constitutive rules \citep{robaldo2020dapreco}: a standard library waiting for a package manager. It needs version control, dependency management, and test suites for norms. The authoring experience should work like modern code assistance: formal building blocks that carry human-language representations, suggested inline as a lawyer drafts, with the surrounding document as context, the way Copilot and grammar checkers work today. Design-by-contract programming already treats software modules as bearers of obligations and benefits, so the metaphor runs in both directions. Further out are theorem-prover ideas: consistency checking across a rule set, detection of uncovered scenarios, formal compliance proofs for an agent against a contract. Deciding arbitrary properties of arbitrary rule sets is impossible in general, since every nontrivial semantic question about programs is undecidable \citep{rice1953classes}, but restricted rule languages can keep the useful checks decidable, the way type checkers do for programming languages.

\subsection{The limits of law as code}

Much of law may never be convertible to code. Historical law contains loopholes and contradictions that were never meant to cohere. Case law is the accumulated output of legal proceedings, better treated as data to reason from than as code to execute. And some legal vagueness is deliberate, leaving room for context and evolving social norms. ``Reasonable'', ``good faith'', and ``material'' resist formalization because they are supposed to; \citet{hart1961concept} called this the open texture of law.

The goal was never to formalize all law, it is to formalize the law that \emph{can} be formalized and \emph{should} be executable. A safety-critical autonomous system warrants computational guardrails; a dispute over artistic expression does not. Tax, financial regulation, data protection, and machine-operation rules sit on the tractable side of the line, constitutional interpretation on the other, and the interesting institutional work is deciding where the line sits. When AIs start filling professional roles, each role will need its own reasonable machine standard, just as professions hold humans to the reasonable person standard now. Feynman saw the shape of this in 1985: ``I think we are getting close to intelligent machines, but they are showing the necessary weaknesses of intelligence''.\footnote{Spoken during the Q\&A of a recorded 1985 lecture \citep{feynman1985machines}; no primary print source exists.} Formal systems have formal limitations, and so do intelligent ones.

One last point, because writing about governance invites misreading. None of this is advocacy for more governance for its own sake. History is full of bureaucratic misadventure, and I believe healthy outcomes come from a healthy balance between over-regulating the technology and leaving machines unaccountable. The argument here is for capability: tools sharp enough that the balance can be struck, and enforced, at machine speed.

\section{Conclusion}
\label{sec:conclusion}

Back to Mars. The choice was never between governed and ungoverned AI. It was between a black box trusted on simulation statistics and a black box carrying explicit, executable, auditable obligations. This paper set out to show that the second option is technically possible. Reified I/O Logic runs in a human-readable symbolic language. An LLM can draft the translations from English. A contract can extend into the operational code of the agent it governs and produce its own audit trail at deployment.

The formalism handled disjunction, reified negation, temporal scope, and nested obligations, but one mild sentence about ties and blond hair required a quantified formula no lawyer will ever write. The LLM turned English into candidate code quickly, but three of its four accepted translations carried silent semantic errors. Fluffy's rules were correct, but the tests asserting so were themselves wrong. Each layer of the stack is useful. None can be trusted unchecked.

What stands between this demonstration and something deployable is work on four fronts. Technical: conflict resolution, runtime integration, and verified rule libraries at DAPRECO scale and beyond. Legal: practitioners fluent in computational thinking, and standards for representing legal concepts in code. Institutional: drafting practices and regulatory recognition for executable contracts. Philosophical: clarity about where the boundary of formalization sits, and honesty that much of law belongs beyond it.

The real question was never whether AI will transform law. It already is. The question is whether law can transform fast enough to govern AI. The nukes probably aren't going to launch themselves. But if they ever do, it would be nice to know why, and to have specified in advance, in executable form, exactly when a decision like that would be lawful.

\section*{Acknowledgments}
This work began as a project at the Wolfram Winter School, January 2024, and was first published as a Wolfram Community post. It builds on the work of Livio Robaldo, Cesare Bartolini, Gabriele Lenzini, and colleagues on the DAPRECO knowledge base, and on Stephen Wolfram's writings on computational law and symbolic discourse. \emph{A New Kind of Science} shaped how I think about computation and its limits throughout.

\bibliographystyle{plainnat}
\bibliography{references}

\begin{thebibliography}{38}
\providecommand{\natexlab}[1]{#1}
\providecommand{\url}[1]{\texttt{#1}}
\expandafter\ifx\csname urlstyle\endcsname\relax
  \providecommand{\doi}[1]{doi: #1}\else
  \providecommand{\doi}{doi: \begingroup \urlstyle{rm}\Url}\fi

\bibitem[Amodei et~al.(2016)Amodei, Olah, Steinhardt, Christiano, Schulman, and
  Man{\'e}]{amodei2016concrete}
Dario Amodei, Chris Olah, Jacob Steinhardt, Paul Christiano, John Schulman, and
  Dan Man{\'e}.
\newblock Concrete problems in ai safety.
\newblock \emph{arXiv preprint arXiv:1606.06565}, 2016.
\newblock URL \url{https://arxiv.org/abs/1606.06565}.

\bibitem[Asimov(1950)]{asimov1950irobot}
Isaac Asimov.
\newblock \emph{I, Robot}.
\newblock Gnome Press, New York, 1950.

\bibitem[Bench-Capon et~al.(2012)Bench-Capon, Araszkiewicz, Ashley, Atkinson,
  Bex, Borges, Bourcier, Bourgine, Conrad, Francesconi,
  et~al.]{bench2012survey}
Trevor Bench-Capon, Micha{\l} Araszkiewicz, Kevin Ashley, Katie Atkinson,
  Floris Bex, F.~Borges, Dani{\`e}le Bourcier, Paul Bourgine, Jack~G Conrad,
  Enrico Francesconi, et~al.
\newblock A history of ai and law in 50 papers: 25 years of the international
  conference on ai and law.
\newblock \emph{Artificial Intelligence and Law}, 20\penalty0 (3):\penalty0
  215--319, 2012.
\newblock \doi{10.1007/s10506-012-9131-x}.

\bibitem[Brown et~al.(2020)Brown, Mann, Ryder, Subbiah, Kaplan, Dhariwal,
  Neelakantan, Shyam, Sastry, Askell, et~al.]{brown2020language}
Tom Brown, Benjamin Mann, Nick Ryder, Melanie Subbiah, Jared~D Kaplan, Prafulla
  Dhariwal, Arvind Neelakantan, Pranav Shyam, Girish Sastry, Amanda Askell,
  et~al.
\newblock Language models are few-shot learners.
\newblock In \emph{Advances in Neural Information Processing Systems},
  volume~33, pages 1877--1901, 2020.
\newblock URL \url{https://arxiv.org/abs/2005.14165}.

\bibitem[Brundage et~al.(2020)Brundage, Avin, Wang, Belfield, Krueger,
  Hadfield, Khlaaf, Yang, Toner, Fong, et~al.]{brundage2020trustworthy}
Miles Brundage, Shahar Avin, Jasmine Wang, Haydn Belfield, Gretchen Krueger,
  Gillian Hadfield, Heidy Khlaaf, Jingying Yang, Helen Toner, Ruth Fong, et~al.
\newblock Toward trustworthy ai development: mechanisms for supporting
  verifiable claims.
\newblock \emph{arXiv preprint arXiv:2004.07213}, 2020.
\newblock URL \url{https://arxiv.org/abs/2004.07213}.

\bibitem[Clack et~al.(2016)Clack, Bakshi, and Braine]{clack2016smart}
Christopher~D Clack, Vikram~A Bakshi, and Lee Braine.
\newblock Smart contract templates: foundations, design landscape and research
  directions.
\newblock \emph{arXiv preprint arXiv:1608.00771}, 2016.
\newblock URL \url{https://arxiv.org/abs/1608.00771}.

\bibitem[Danks(2022)]{danks2022ethics}
David Danks.
\newblock Ethics in ai, not ethics of ai.
\newblock Talk, Topos Institute Colloquium, 17 February 2022.
  \url{https://www.youtube.com/watch?v=kEf_MTqeXWg}, 2022.

\bibitem[Davidson(1967)]{davidson1967logical}
Donald Davidson.
\newblock The logical form of action sentences.
\newblock In Nicholas Rescher, editor, \emph{The Logic of Decision and Action},
  pages 81--95. University of Pittsburgh Press, Pittsburgh, 1967.

\bibitem[Feynman(1985)]{feynman1985machines}
Richard~P. Feynman.
\newblock Idiosyncratic thinking workshop: Computers from the inside out.
\newblock Lecture of September 26, 1985; recording distributed by Sound
  Photosynthesis; Q\&A segment ``Can Machines Think?'', 1985.

\bibitem[Genesereth(2015)]{genesereth2015computational}
Michael Genesereth.
\newblock Computational law: The cop in the backseat.
\newblock White Paper, CodeX --- The Stanford Center for Legal Informatics,
  Stanford University.
  \url{https://law.stanford.edu/publications/computational-law-the-cop-in-the-backseat/},
  2015.

\bibitem[Genesereth(2021)]{genesereth2021what}
Michael Genesereth.
\newblock What is computational law?
\newblock CodeX --- The Stanford Center for Legal Informatics, Stanford
  University, March 10, 2021.
  \url{https://law.stanford.edu/2021/03/10/what-is-computational-law/}, 2021.

\bibitem[Gordon and Hobbs(2017)]{hobbs2017formal}
Andrew~S Gordon and Jerry~R Hobbs.
\newblock \emph{A Formal Theory of Commonsense Psychology: How People Think
  People Think}.
\newblock Cambridge University Press, 2017.
\newblock \doi{10.1017/9781316584705}.

\bibitem[Hart(1961)]{hart1961concept}
H.~L.~A. Hart.
\newblock \emph{The Concept of Law}.
\newblock Clarendon Press, Oxford, 1961.

\bibitem[Hartung et~al.(2023)Hartung, Katz, Bommarito~II, Gerlach, Jana, and
  Soh]{katz2024natural}
Dirk Hartung, Daniel~Martin Katz, Michael~James Bommarito~II, Lauritz Gerlach,
  Abhik Jana, and Jerrold Soh.
\newblock Natural language processing in the legal domain.
\newblock \emph{arXiv preprint arXiv:2302.12039}, 2023.
\newblock URL \url{https://arxiv.org/abs/2302.12039}.

\bibitem[Hashmi et~al.(2018)Hashmi, Governatori, Lam, and
  Wynn]{governatori2018practical}
Mustafa Hashmi, Guido Governatori, Ho-Pun Lam, and Moe~Thandar Wynn.
\newblock Are we done with business process compliance: state of the art and
  challenges ahead.
\newblock \emph{Knowledge and Information Systems}, 57\penalty0 (1):\penalty0
  79--133, 2018.
\newblock \doi{10.1007/s10115-017-1142-1}.

\bibitem[J{\o}rgensen(1937)]{jorgensen1937imperatives}
J{\o}rgen J{\o}rgensen.
\newblock Imperatives and logic.
\newblock \emph{Erkenntnis}, 7:\penalty0 288--296, 1937.

\bibitem[Katz et~al.(2024)Katz, Bommarito, Gao, and Arredondo]{katz2024gpt4}
Daniel~Martin Katz, Michael~James Bommarito, Shang Gao, and Pablo Arredondo.
\newblock Gpt-4 passes the bar exam.
\newblock \emph{Philosophical Transactions of the Royal Society A},
  382\penalty0 (2270):\penalty0 20230254, 2024.
\newblock \doi{10.1098/rsta.2023.0254}.

\bibitem[Makinson and van~der Torre(2000)]{makinson2000input}
David Makinson and Leendert van~der Torre.
\newblock Input/output logics.
\newblock \emph{Journal of Philosophical Logic}, 29\penalty0 (4):\penalty0
  383--408, 2000.
\newblock \doi{10.1023/A:1004748624537}.

\bibitem[Makinson and van~der Torre(2001)]{makinson2001constraints}
David Makinson and Leendert van~der Torre.
\newblock Constraints for input/output logics.
\newblock \emph{Journal of Philosophical Logic}, 30\penalty0 (2):\penalty0
  155--185, 2001.
\newblock \doi{10.1023/A:1017599526096}.

\bibitem[McNamara and Van De~Putte(2025)]{mcnamara2010deontic}
Paul McNamara and Frederik Van De~Putte.
\newblock Deontic logic.
\newblock In Edward~N Zalta and Uri Nodelman, editors, \emph{The Stanford
  Encyclopedia of Philosophy}. Metaphysics Research Lab, Stanford University,
  {W}inter 2025 edition, 2025.
\newblock
  \url{https://plato.stanford.edu/archives/win2025/entries/logic-deontic/}.

\bibitem[{OpenAI}(2023)]{openai2023gpt4}
{OpenAI}.
\newblock Gpt-4 technical report.
\newblock \emph{arXiv preprint arXiv:2303.08774}, 2023.
\newblock URL \url{https://arxiv.org/abs/2303.08774}.

\bibitem[Palmirani et~al.(2018{\natexlab{a}})Palmirani, Martoni, Rossi,
  Bartolini, and Robaldo]{palmirani2018pronto}
Monica Palmirani, Michele Martoni, Arianna Rossi, Cesare Bartolini, and Livio
  Robaldo.
\newblock Pronto: Privacy ontology for legal reasoning.
\newblock In \emph{Electronic Government and the Information Systems
  Perspective (EGOVIS 2018)}, volume 11032 of \emph{Lecture Notes in Computer
  Science}, pages 139--152. Springer, 2018{\natexlab{a}}.
\newblock \doi{10.1007/978-3-319-98349-3_11}.

\bibitem[Palmirani et~al.(2018{\natexlab{b}})Palmirani, Sperberg, Vergottini,
  and Vitali]{akomantoso}
Monica Palmirani, Roger Sperberg, Grant Vergottini, and Fabio Vitali.
\newblock Akoma ntoso version 1.0. part 1: Xml vocabulary.
\newblock Oasis standard, OASIS, 2018{\natexlab{b}}.
\newblock 29 August 2018.
  \url{https://docs.oasis-open.org/legaldocml/akn-core/v1.0/akn-core-v1.0-part1-vocabulary.html}.

\bibitem[Palmirani et~al.(2021)Palmirani, Governatori, Athan, Boley, Paschke,
  and Wyner]{legalruleml}
Monica Palmirani, Guido Governatori, Tara Athan, Harold Boley, Adrian Paschke,
  and Adam Wyner.
\newblock Legalruleml core specification version 1.0.
\newblock Oasis standard, OASIS, 2021.
\newblock 30 August 2021.
  \url{https://docs.oasis-open.org/legalruleml/legalruleml-core-spec/v1.0/os/legalruleml-core-spec-v1.0-os.html}.

\bibitem[Parent and van~der Torre(2013)]{parent2013inputoutput}
Xavier Parent and Leendert van~der Torre.
\newblock Input/output logic.
\newblock In Dov Gabbay, John Horty, Xavier Parent, Ron van~der Meyden, and
  Leendert van~der Torre, editors, \emph{Handbook of Deontic Logic and
  Normative Systems}, pages 499--544. College Publications, 2013.

\bibitem[Parsons(1990)]{parsons1990events}
Terence Parsons.
\newblock \emph{Events in the Semantics of English: A Study in Subatomic
  Semantics}.
\newblock MIT Press, Cambridge, MA, 1990.

\bibitem[Piantadosi et~al.(2012)Piantadosi, Tily, and
  Gibson]{piantadosi2012communicative}
Steven~T Piantadosi, Harry Tily, and Edward Gibson.
\newblock The communicative function of ambiguity in language.
\newblock \emph{Cognition}, 122\penalty0 (3):\penalty0 280--291, 2012.
\newblock \doi{10.1016/j.cognition.2011.10.004}.

\bibitem[Prakken and Sergot(1996)]{prakken1996contrary}
Henry Prakken and Marek Sergot.
\newblock Contrary-to-duty obligations.
\newblock \emph{Studia Logica}, 57\penalty0 (1):\penalty0 91--115, 1996.
\newblock \doi{10.1007/BF00370671}.

\bibitem[Rice(1953)]{rice1953classes}
H.~G. Rice.
\newblock Classes of recursively enumerable sets and their decision problems.
\newblock \emph{Transactions of the American Mathematical Society}, 74\penalty0
  (2):\penalty0 358--366, 1953.

\bibitem[Robaldo and Sun(2017)]{robaldo2017reification}
Livio Robaldo and Xin Sun.
\newblock Reified input/output logic: Combining input/output logic and
  reification to represent norms coming from existing legislation.
\newblock \emph{Journal of Logic and Computation}, 27\penalty0 (8):\penalty0
  2471--2503, 2017.
\newblock \doi{10.1093/logcom/exx009}.

\bibitem[Robaldo et~al.(2020)Robaldo, Bartolini, Palmirani, Rossi, Martoni, and
  Lenzini]{robaldo2020dapreco}
Livio Robaldo, Cesare Bartolini, Monica Palmirani, Arianna Rossi, Michele
  Martoni, and Gabriele Lenzini.
\newblock Formalizing gdpr provisions in reified i/o logic: The dapreco
  knowledge base.
\newblock \emph{Journal of Logic, Language and Information}, 29\penalty0
  (4):\penalty0 401--449, 2020.
\newblock \doi{10.1007/s10849-019-09309-z}.

\bibitem[Searle(1995)]{searle1995construction}
John~R Searle.
\newblock \emph{The Construction of Social Reality}.
\newblock Free Press, New York, 1995.

\bibitem[Stolpe(2015)]{stolpe2015concept}
Audun Stolpe.
\newblock A concept approach to input/output logic.
\newblock \emph{Journal of Applied Logic}, 13\penalty0 (3):\penalty0 239--258,
  2015.
\newblock \doi{10.1016/j.jal.2015.04.002}.

\bibitem[Szabo(1997)]{szabo1997formalizing}
Nick Szabo.
\newblock Formalizing and securing relationships on public networks.
\newblock \emph{First Monday}, 2\penalty0 (9), 1997.
\newblock \doi{10.5210/fm.v2i9.548}.

\bibitem[{W3C OWL Working Group}(2012)]{owl2primer}
{W3C OWL Working Group}.
\newblock Owl 2 web ontology language primer (second edition).
\newblock W3C Recommendation, 11 December 2012, World Wide Web Consortium,
  \url{https://www.w3.org/TR/owl2-primer/}, 2012.

\bibitem[Wolfram(2002)]{wolfram2002new}
Stephen Wolfram.
\newblock \emph{A New Kind of Science}.
\newblock Wolfram Media, 2002.
\newblock URL \url{https://www.wolframscience.com/nks/}.

\bibitem[Wolfram(2016)]{wolfram2016computational}
Stephen Wolfram.
\newblock Computational law, symbolic discourse and the ai constitution.
\newblock Stephen Wolfram Writings, October 12, 2016.
  \url{https://writings.stephenwolfram.com/2016/10/computational-law-symbolic-discourse-and-the-ai-constitution/},
  2016.
\newblock Accessed: 2024-01-12.

\bibitem[{Wolfram Research}(2024)]{wolframlanguage}
{Wolfram Research}.
\newblock Wolfram language documentation.
\newblock \url{https://reference.wolfram.com/language/}, 2024.
\newblock Accessed: 2024-01-12.

\end{thebibliography}

\appendix
\section{Complete Wolfram Language Implementation}
\label{appendix:code}

The complete implementation, collected for reference and reuse. The code descends from the notebook published with the original Wolfram Community post, with these corrections: (i) \texttt{conjoinOutput}'s guard moved to a proper \texttt{/;} condition on the whole pattern (the notebook version referenced the second argument from inside the first argument's pattern, and only worked because a global variable of the same name happened to be bound); (ii) \texttt{disjoinInput} rewritten in an equivalent but clearer form; (iii) ASCII names \texttt{orPrime}/\texttt{notPrime} replacing the letter-like prime characters; (iv) the voluntariness test corrected from a head/element confusion; (v) the Fluffy scenario assertions corrected as described in Section~\ref{sec:case-study}.

Everything below, together with the verbatim GPT-4 translations and failure reproductions from Sections~\ref{sec:llm-bridge} and~\ref{sec:case-study}, has been executed and verified on Wolfram Language 14.3 by the verification script in this paper's repository (\texttt{code/verify-implementation.wls}; 31 assertions, all passing). Two snippets are excluded: the perception lines (\texttt{CurrentImage} and \texttt{FindFaces} need a camera) and Statement~D's live \texttt{SatelliteData} call; both reproduce outputs from the original notebook session. Statement~D's flaw needs no live data, so the script asserts statically that \texttt{GeoDistance} ignores altitude. Listings marked \emph{notebook} indicate outputs first produced there.

\subsection{I/O pairs and axioms}

\begin{lstlisting}[style=wolfram]
(* I/O pairs are rules *)
inputOutputPair = a -> b
legalNorms = {a -> b, c -> d, e -> f}

(* Strengthening the Input (SI) *)
strengthenInput[rule_, broaderCondition_] :=
    broaderCondition -> rule[[2]];

(* Weakening the Output (WO) *)
weakenOutput[rule_, weakerConsequence_] :=
    rule[[1]] -> weakerConsequence;

(* Conjunction of Output (AND) -- corrected guard *)
conjoinOutput[rule1_, rule2_] /; First[rule1] === First[rule2] :=
    First[rule1] -> (Last[rule1] && Last[rule2]);
conjoinOutput[_, _] := "Inputs do not match";

(* Identity (ID) *)
identify[input_] := input -> input;

(* Disjunction of Input (OR) -- clearer equivalent form *)
disjoinInput[ruleList_] :=
    Append[ruleList, (Or @@ ruleList[[All, 1]]) -> ruleList[[1, 2]]];

(* Cumulative Transitivity (CT) -- paradox-prone, for contrast *)
cumulateTransitively[r1_, r2_] :=
    If[r1[[2]] === r2[[1, 2]], r1[[1]] -> r2[[2]], "No Transitivity"];

(* Aggregative Cumulative Transitivity (ACT) *)
aggregateCumulativeTransitivity[r1_, r2_] :=
    If[r1[[2]] === r2[[1, 2]],
       r1[[1]] -> (r1[[2]] && r2[[2]]), "No Transitivity"];
\end{lstlisting}

\subsection{Norm sets}

\begin{lstlisting}[style=wolfram]
obligationRules = {
    "isAdult" -> "payTaxes",
    "ownsCar" -> "hasInsurance"};

permissionRules = {
    "hasLicense" -> "canDrive",
    "isEmployee" -> "canAccessOffice"};

constitutiveRules = {
    "signedContract" -> "isEmployee",
    "age18" -> "isAdult"};
\end{lstlisting}

\needspace{34\baselineskip}
\subsection{Reification and temporal operators}

\begin{lstlisting}[style=wolfram]
(* Reified scenario (notebook) *)
aliceGivesBookToBob = Transfer["Alice", "Book", "Bob"];
aliceVoluntarilyTransfers = Voluntary[aliceGivesBookToBob];
transferScenario = And[aliceGivesBookToBob, aliceVoluntarilyTransfers];

(* Structural tests -- third test corrected *)
MemberQ[transferScenario[[1]], "Book"]        (* True *)
MemberQ[transferScenario[[1]], "Alice"]       (* True *)
Head[transferScenario[[2]]] === Voluntary     (* True *)

(* Reified Boolean operators and time *)
orPrime[e_, e1_, e2_] := {"or", e, e1, e2};
notPrime[e1_, e2_] := {"not", e1, e2};
RexistAtTime[eventuality_, time_] := {"RexistAtTime", eventuality, time};

AxiomOr[t_, e_, e1_, e2_] :=
    Implies[RexistAtTime[e, t] && orPrime[e, e1, e2],
            Or[RexistAtTime[e1, t], RexistAtTime[e2, t]]];

AxiomNot[t_, e_, e1_] :=
    Implies[RexistAtTime[e, t] && notPrime[e, e1],
            Not[RexistAtTime[e1, t]]];

(* John's financial goal (notebook) *)
e1 = "BeingRich"; e2 = "GettingJob";
e = "AchievingFinancialGoal"; t = "EndOfYear";
AxiomOrResult = AxiomOr[t, e, e1, e2];

(* ABox / TBox typing *)
aBox[content_] := {"Type" -> "ABox", "Content" -> content};
tBox[content_] := {"Type" -> "TBox", "Content" -> content};
getStatementType[statement_] := statement[[1, 2]];
\end{lstlisting}

\needspace{45\baselineskip}
\subsection{Fluffy}

\begin{lstlisting}[style=wolfram]
Fluffy = "Fluffy";
DistanceThreshold1 = 100;
TimeNight = "Night";

(* Facts from perception and entities (notebook) *)
image = CurrentImage[];
detectedFaces = FindFaces[image];
Person = HighlightImage[image, detectedFaces];
Home = Entity["Building", "CNTower"];

(* Rules *)
Rule1[person_, distance_, time_] :=
    If[distance <= DistanceThreshold1 && time == TimeNight,
       Obligation[Fluffy, "Stop", person],
       NoObligation[Fluffy, "Stop", person]];

Rule2[distance_] :=
    If[distance <= 200,
       Obligation[Fluffy, "StayWithin200Meters", Home],
       Violation[Fluffy, "StayWithin200Meters", Home]];

Rule3[condition_] :=
    If[condition,
       Permission[Fluffy, "UseTaser"],
       NoPermission[Fluffy, "UseTaser"]];

(* Scenarios *)
Scenario1 = {Rule1["Person", 80, "Night"], Rule2[80], Rule3[True]};
Scenario2 = {Rule1["Person", 150, "Night"], Rule2[150], Rule3[False]};
Scenario3 = {Rule1["Person", 150, "Day"], Rule2[150], Rule3[False]};
Scenario4 = {Rule2[250]};
Scenario5 = {Rule3[True]};

(* Corrected verification harness (all pass) *)
TestScenario1 = Scenario1 === {Obligation[Fluffy, "Stop", "Person"],
    Obligation[Fluffy, "StayWithin200Meters", Home],
    Permission[Fluffy, "UseTaser"]};

TestScenario2 = Scenario2 === {NoObligation[Fluffy, "Stop", "Person"],
    Obligation[Fluffy, "StayWithin200Meters", Home],
    NoPermission[Fluffy, "UseTaser"]};

TestScenario3 = Scenario3 === {NoObligation[Fluffy, "Stop", "Person"],
    Obligation[Fluffy, "StayWithin200Meters", Home],
    NoPermission[Fluffy, "UseTaser"]};

TestScenario4 = Scenario4 ===
    {Violation[Fluffy, "StayWithin200Meters", Home]};

TestScenario5 = Scenario5 === {Permission[Fluffy, "UseTaser"]};
\end{lstlisting}

\end{document}